\pdfoutput=1

\documentclass[11pt]{article}

\usepackage[]{acl}

\usepackage{times}
\usepackage{latexsym}
\usepackage{graphicx}
\usepackage{amsfonts}

\usepackage[T1]{fontenc}

\usepackage[utf8]{inputenc}

\usepackage{microtype}

\definecolor{ao(english)}{rgb}{0.0, 0.5, 0.0}
\definecolor{applegreen}{rgb}{0.55, 0.71, 0.0}
\definecolor{bananayellow}{rgb}{1.0, 0.88, 0.21}

\title{LOViS: \textbf{L}earning \textbf{O}rientation and \textbf{Vi}sual \textbf{S}ignals for \\
Vision and Language Navigation}

\author{Yue Zhang \\
  Michigan State University\\
  \texttt{zhan1624@msu.edu} \And
  Parisa Kordjamshidi \\
  Michigan State University\\
  \texttt{kordjams@msu.edu} \\}

\begin{document}
\maketitle

\begin{abstract}

Understanding spatial and visual information is essential for a navigation agent who follows natural language instructions.
The current Transformer-based VLN agents entangle the orientation and vision information, which limits the gain from the learning of each information source. In this paper, we design a neural agent with explicit Orientation and Vision modules. Those modules learn to ground spatial information and landmark mentions in the instructions to the visual environment more effectively. 
To strengthen the spatial reasoning and visual perception of the agent, we design specific pre-training tasks to feed and better utilize the corresponding modules in our final navigation model.
We evaluate our approach on both Room2room (R2R) and Room4room (R4R) datasets and achieve the state of the art results on both benchmarks.
\end{abstract}
\section{Introduction}
Vision and Language Navigation (VLN)  problem~\cite{anderson2018vision} has attracted increasing attention from the communities of computer vision, natural language processing, and robotics because of its broad real-world applications. In this problem setting, the goal of a navigation agent is to move to a target location in a photo-realistic simulated environment by following a detailed natural language instruction, e.g., \textit{``Walk into the bedroom. Walk past the bedroom door. Wait at the laundry room door.''}. 
Two kinds of simulators are used to create the dataset and the corresponding problem formulation: discrete trajectories~\cite{anderson2018vision} and continuous navigation trajectories~\cite{krantz2020beyond}. In this paper, we work on the discrete one, that an agent traverses a pre-defined connectivity graph by selecting the adjacent viewpoint with a higher probability of corresponding to the instruction at each step. 
\begin{figure}[!t]
\centering
\includegraphics[width=1.\linewidth]{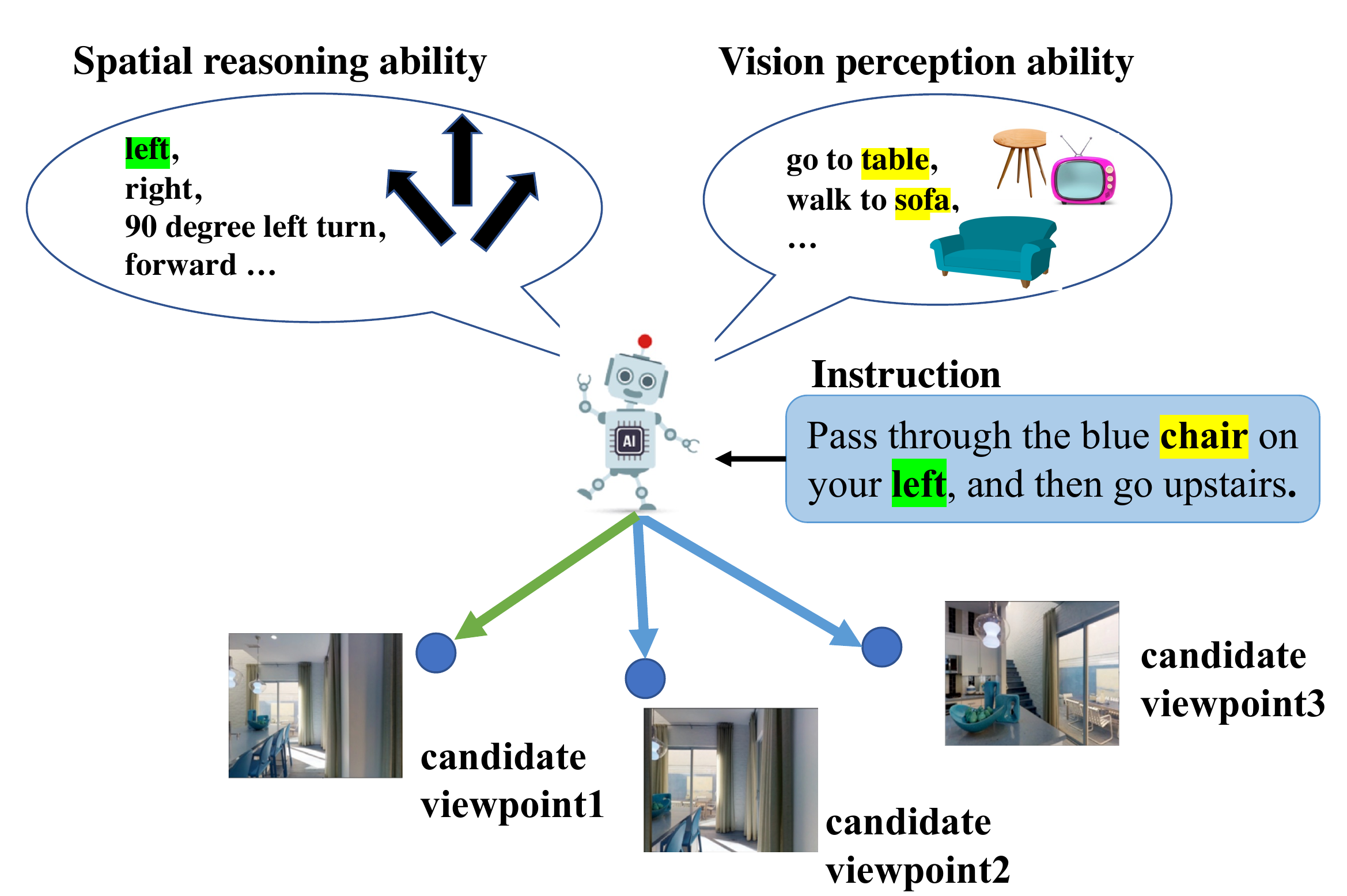}
\caption{\small 
Spatial reasoning helps leveraging orientation clues, such as \textit{left} and \textit{90 degree}; visual perception ability grounds mentioned landmarks, such as \textit{table}, \textit{sofa}, and \textit{chair}. With these two abilities,
the agent selects the
candidate viewpoint corresponding to the instruction
at each navigation step. The green arrow shows the ground-truth viewpoint. \vspace{-7mm}
}
\label{fig: demonstration of VLN task}
\end{figure}

The earlier research in the VLN area can be divided into two categories. The first category of models mostly depends on the encoder-decoder framework for encoding the text and visual information, establishing the connections between them with the attention mechanism, and decoding the actions~\cite{anderson2018vision,ma2019self,fried2018speaker}. 
The second category of works learns to model the semantic structure which implicitly or explicitly enhances the textual-visual matching~\cite{hong2020language, hong2020sub, qi2020object, zhang2021towards,li2021improving, zhang2022explicit}. However, these prior research are surpassed by the most recently proposed Transformer-based VLN agents~\cite{hong2021vln, hao2020towards,guhur2021airbert, chen2021history} which capture the cross-modality information and demonstrate an outstanding navigation performance. 

As shown in the example of Figure~\ref{fig: demonstration of VLN task}, two major abilities are important to the navigation agent: 
\textit{spatial reasoning} and \textit{visual perception}. 
While navigation seems to require both of these abilities, there are cases where understanding even one of these is sufficient.
For example, \textit{spatial reasoning} guides the agent towards the correct direction when the instruction is ``\textit{90-degree left-turn}'' or ``\textit{on your right}'', regardless of the surrounding visual scene or objects. In some other cases, \textit{visual perception} is sufficient to recognize the mentioned landmarks in the visual environment after receiving instructions without any auxiliary signals of orientation, such as ``\textit{walk to the sofa}'' or  ``\textit{pass the table}''.


The current Transformer-based agents tend to intertwine the learning of these two abilities, which we argue may impede developing a more effective navigation model. In the light of this, we propose a new navigation agent with different modules to select actions based on orientation and vision perspectives separately. Moreover,
we design specific pre-training tasks to distil more explicit spatial and visual knowledge, which is better utilized in the corresponding modules in our navigation agent.
This is different from the majority of methods employing pre-training tasks without considering the needs of the target downstream tasks. Our modular design interacts with modular pre-training, guiding the agents to generate specialized features which can be better adapted to the downstream tasks.

Specifically, in the downstream navigation task, in order to utilize the learnt spatial and visual information, we design a framework, called LOViS. It contains three modules, namely, \textit{history module}, \textit{orientation module}, and \textit{vision module}. In the \textit{history module,} the agent uses the previous step's information to determine which tokens in the instruction it should pay attention to. And then, the agent learns to connect such attended tokens to the corresponding visual information
to finally make a history-based action decision. In the \textit{orientation module}, the agent only focuses on orientation information in the instruction (\textit{e.g.,} \textit{left} and $\textit{90}$ \textit{degree}), and then grounds them to the vision environment to make an orientation-based action. Likewise, in the \textit{vision module}, the agent is encouraged to focus on the mentioned landmarks in the instruction (\textit{e.g.,} \textit{table}, \textit{lamp} and \textit{chair}), and ground them into the vision to obtain a vision-based action decision. Finally, the agent combines the action decisions of three modules to make the final decision.

In the pre-training process,  we propose two specific pre-training tasks, namely \textit{Orientation Matching} (OM) and \textit{Vision Matching} (VM), to learn orientation and vision information, respectively. Besides, we modify two commonly used pre-training tasks for navigation: Masked Language Modeling (MLM) and Single Step Action Prediction (SSAP), to obtain better cross-modal representations for the downstream navigation model.



In summary, our contributions are as follows:
\\
1. Unlike previous models, our novel Transformer-based agent includes \textbf{two new modules to capture the orientation and visual information signals separately}. This benefits the agent from both of these information sources to select an action more effectively. \\
2. We design \textbf{new pre-training tasks} to emphasize (a) learning spatial reasoning and grounding the orientation information in the environment; (b) learning visual perception and grounding landmark mentions in the environment. These pre-training representations are utilized in the corresponding modules in the navigation model.\\
3. Our method \textbf{exceeds the current SOTA} results on both \textit{Room2room} and \textit{Room4Room} benchmarks.


\section{Related Work}
\begin{figure*}[!t]
    \centering
    \includegraphics[height=7.0cm, width=1.\linewidth]{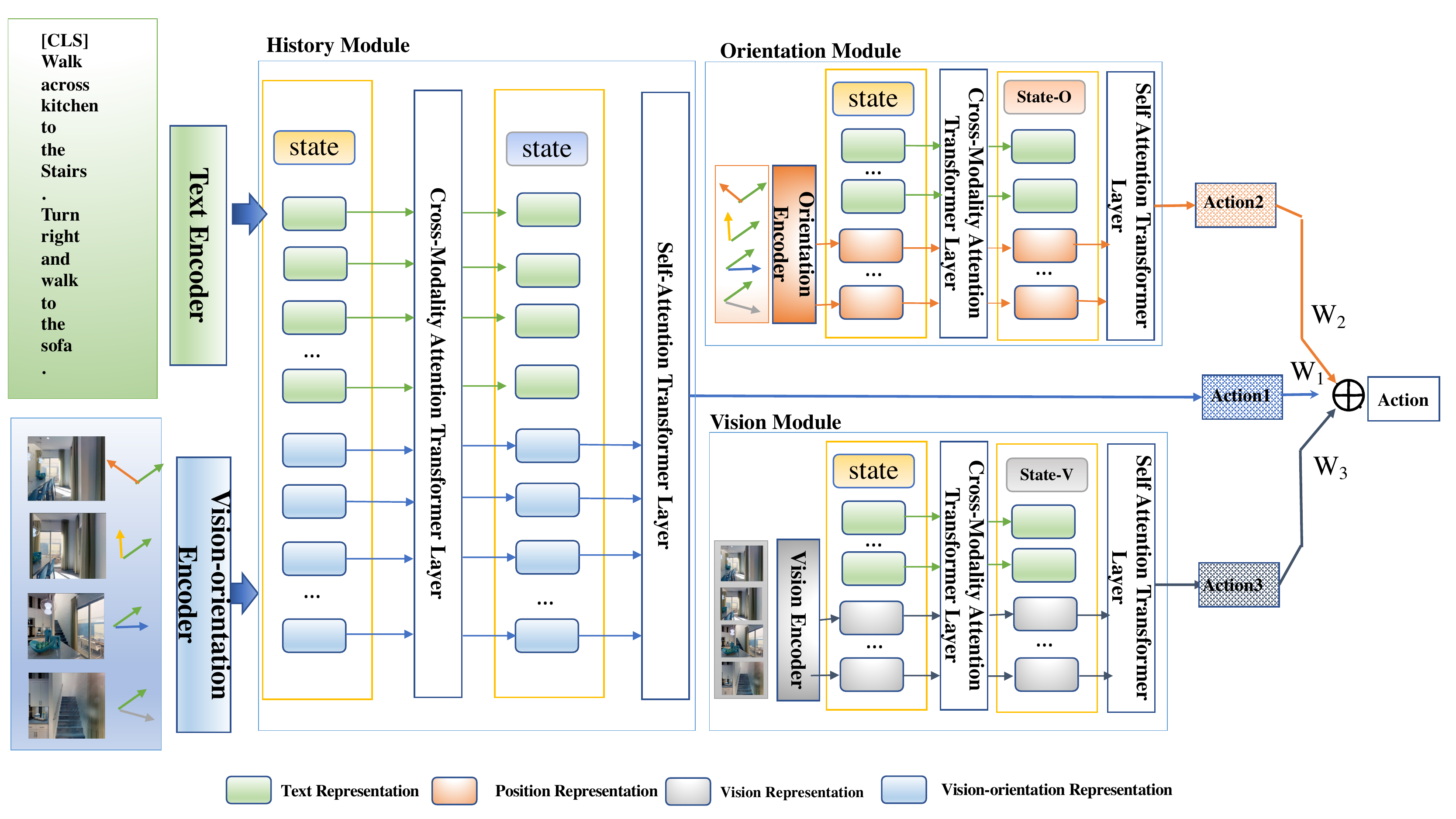}
    \caption{\small Our proposed LOViS contains three modules: history module, orientation module and vision module. Each module can generate action decision based on different reasoning, then three actions are combined to determine the final action selection. \vspace{-4mm}
    }
    \label{fig:model architecture}
\end{figure*}
\textbf{Vision-and-Language Navigation} 
Many deep learning methods~\cite{tan2019learning, hong2021vln, hong2020language} for VLN tasks have been proposed in the past few years.
For example, ~\newcite{anderson2018vision} firstly proposed a Sequence-to-Sequence baseline model to encode the instructions and decode the embeddings to the low-level action sequence with the observed images. SF~\cite{fried2018speaker} generates the augmented samples to address the generalization issues and extends the low-level space to panoramic action space. RelGraph~\cite{hong2020language} builds an implicit language-visual entity relation graph to learn the connection between the text and vision modalities. EXOR~\cite{zhang2022explicit} first splits the long instructions into spatial configurations~\cite{dan2020spatial, zhang2021towards, kordjamshidi2010spatial}. Then they explicitly align the landmarks and spatial relations in the spatial configuration to the corresponding information in the visual modality. 
In terms of learning spatial and vision information separately,
OAAM~\cite{qi2020object} is the earliest attempt that decomposes the instruction into action and object phrases and related them to the visual environment to make the final decisions.
NvEM~\cite{an2021neighbor} extends OAAM to divide object module to subject and reference modules and fuse the information from the neighbor views. 
However, to our best knowledge, there is no work investigating how to model spatial and visual information in the Transformer-based VLN agent.\\
\textbf{Transformer-based Navigation Agent} Compared with conventional methods, the Transformer-based model in VL tasks show great improvements~\cite{tan2019lxmert, chen2019uniter, lu2019vilbert, li2020oscar}. However, different from conventional VL tasks (\textit{i.e.,} image captioning), the VLN task requires learned representation to facilitate the action selection, which is a Markov Decision Process. Therefore, the VLN needs to learn the correspondence between language and dynamic visual observation by interacting with the environment.
In the past few years, the VLN task has been formulated as a dynamic grounding problem between texts and images. 
PRESS~\cite{li2021improving} firstly fine-tunes a pre-trained language model BERT to obtain the text representation.
PREVALENT~\cite{hao2020towards} trains a VL Transformer with a large amount of image-text-action triplets to learn cross representations for the navigation task. RecBERT~\cite{hong2021vln} designs a state unit to store history information and train Transformer recurrently for the direct navigation. HAMT~\cite{chen2021history} proposes to explicitly encode  all past observations and actions as history. Also, they improve the performance by changing the fixed vision features to the Vision Transformer,  ViT~\cite{dosovitskiy2020image}. However, all of those transformer-based models entangle the learning of spatial information and visual information.
Furthermore, prior works~\cite{chen2021history,qiao2022hop} design the pre-training tasks without considering the adaption to the downstream model. Our work designs different modules to better utilize the learnt spatial and visual information from the pre-training.
\section{Method}

\subsection{Problem Description}
In this task, formally, the agent is given an instruction $ \mathcal{W} =\{w_1, w_2, \cdots, w_L \}$, where $w$ represents tokens and $L$ is the number of tokens. 
At each time step $t$, the agent observes a panorama which consists of 36 discrete images~\footnote{12 headings and 3 elevations with 30 degree intervals.}, which are denoted as $\mathcal{I}^p = \{I^p_1,I^p_2, \cdots, I^p_{36}\}$. There are $k$ navigable viewpoints in those panoramic views that the agent can navigate to. We denote the navigable viewpoints as $\mathcal{I}^c = \{I^c_1,I^c_2, \cdots, I^c_k\}$.  In our model, to focus on the relevant observations in the visual environment, we only use the navigable viewpoints rather than all 36 images in the whole panoramic view.

The goal of the task is to select the next viewpoint among navigable viewpoints 
which forms a trajectory 
that takes the agent close to a goal destination. The agent terminates when the current viewpoint is selected or a pre-defined maximum number of navigation steps have been reached.

%

\subsection{Background}
\label{background}
Following~\cite{hong2021vln}, 
we design text encoder, vision features, and a recurrent state unit.\\
\textbf{Text Encoder} We first apply BERT tokenizer to split the text instruction to a sequence of tokens: \{$[CLS], w_1, w_2, \cdots, w_L, [SEP]$\}, where $L$ is the number of tokens, and $[CLS]$ and $[SEP]$ are the special tokens.
Text embedding of each token is obtained by summing up the token embedding, position embedding and type embedding of text. Then the text embedding is passed through a text encoder, a standard multi-layer Transformer with self-attention, to obtain the contextual representation, represented as $X=\{x_1, x_2, \cdots, x_L\}$.\\
\textbf{Vision Features}
For each navigable viewpoint, we consider its vision and relative orientation features.
Specifically, we use ResNet-152~\cite{he2016deep} pre-trained on ImageNet~\cite{russakovsky2015imagenet} as 2048-d vision representation. We repeat relative heading $\alpha$ and elevation $\beta$ features $[sin \alpha; cos\alpha; sin\beta, cos\beta]$ 32 times to constitute a 128-d orientation feature as $O_p$.
Formally, we denote the vision and orientation representations for the the navigable viewpoints as $V = \{v_1, v_2, \cdots, v_k\}$ and $O = \{o_1, o_2, \cdots, o_k\}$, respectively.\\
\textbf{Recurrent State Unit}
The recurrent state unit stores the history information from the previous steps. At each time step, the navigation agent takes three inputs: the state representation $s_{t}$, the language representation $X$, and the vision representation $V_t$. For the next navigation step, the state is refined using the text information and the current observations in the visual environment as follows,
\begin{equation}
    \small
    \vspace{-2mm}
    s_{t+1} = NAV(s_{t}, X, V_t),
\end{equation}
where $NAV$ is the navigation model. In three modules from our model, state representation are initialized with the $[CLS]$ representation in the text encoder. In our model, we assign the state representation to three different modules to consider different information.

\subsection{Our Model: LOViS}
Our proposed model (LOViS) has three main modules: history module, orientation module, and vision module, as depicted in Figure~\ref{fig:model architecture}. \\
\textbf{History Module}
The History Module receives three types of inputs: 
state representation $s_t$ (see ``state'' in Figure~\ref{fig:model architecture}), text representation $X$, and ``vision-orientation'' representations. 
To obtain ``vision-orientation'' representation, we feed the concatenation of vision and orientation representations to a
``vision-orientation encoder" (see Figure~\ref{fig:model architecture}). We denote ``vision-orientation'' representation as $\tilde{VO} = \{\tilde{vo}_1, \tilde{vo}_2, \cdots, \tilde{vo}_k\}$.
Then we use cross-modal attention layers and self-attention layers to obtain the cross representation. In cross-modality attention Transformer layers, one modality is used as a query and the other as the key to exchange information as follows,
\begin{equation}
    \small
    \hat{X}, \hat{s}_t, \hat{VO}_t= Cross\_Attn(X,[s_t; VO_t]),
\end{equation}
where $\hat{X}$, $\hat{s}_t$, and $\hat{VO}_t$ are respectively updated state, text and ``vision-orientation'' representations after cross modality attention layers. Then state and ``vision-orientation'' representations are fed into self-attention Transformer layers: 
\begin{equation}
    \small
    s_{t+1}, p^h_t= Self\_Attn([\hat{s}_t; \hat{VO}_t])\\
    \vspace{-1mm}
\end{equation}
where $s_{t+1}$ is the updated state after self-attention layers. $p^h_t$ is the self attention score between state representations and ``vision-orientation'' representations. 
Note that the refinement of the state representation only happens in the history module.
\\
\textbf{Orientation Module} 
Orientation information is vital for the navigation task. For example, the instruction, ``\textit{turn left}" can assist the agent to ignore the navigable viewpoints on the right side.
In our work, we build an orientation module specifically to encourage the agent to learn the spatial information from the instructions and ground it in the visual environment.
Specifically, we linearly project the orientation features $O$ (refer to the notations in Section~\ref{background}) via the ``Orientation Encoder'' (see Figure~\ref{fig:model architecture}) to obtain its projected representation, denoted as $\tilde{O}$. Then we input the state representation $s_t$, text representation $X$, and the projected orientation representation $\tilde{O}$ to the cross-modality attention Transformer layer. The orientation module learns a new state representation, denoted as $s^o_t$, for orientation information (see ``State-O' in Figure~\ref{fig:model architecture}). 
For cross-model attention layers, we have:  
\begin{equation}
    \vspace{-2mm}
    \small
    \hat{X^o}, \hat{s}^o_t, \hat{O}_t= Cross\_Attn(X,[s^o_t; \tilde{O}_t])
\end{equation}
where $\hat{X^o}, \hat{s}^o_t, \hat{O}_t$ are updated state, text, orientation representations after cross modality attention layers in the orientation module.
Then we use the state representation enriched with the orientation information to perform self-attention with orientation representations as follows.
\begin{equation}
    \small
    p^o_t= Self\_Attn([\hat{s}^o_t; \hat{O}_t])\\
\end{equation}
where $p^o_t$ is the attention score between state representation and orientation feature.\\
\textbf{Vision Module} 
Connecting mentioned landmarks in the instruction to the scene and objects in the visual environment is also important to the navigation task. In the instruction, ``\textit{enter into the bedroom and move close to TV.}'', The mentioned landmarks, such as ``\textit{bedroom}'' and ``\textit{TV}'', provide apparent clues for the navigation actions. Like the orientation module, we build a vision module to ground the text landmarks in the visual scene and objects.
Specifically, we first project vision representations $V$ (refer to the notations in Section~\ref{background}) using ``Vision Encoder''~(see Figure~\ref{fig:model architecture}) to obtain the projected visual representation, denoted as $\tilde{V}$. Then we input the state representation $s_t$, text representation $X$, and projected vision representation $\tilde{V}$ to the cross-modal attention and self-attention layers as follows,
\begin{equation}
    \small
    \hat{X}^v, \hat{s}^v_t, \hat{V}_t= Cross\_Attn(X,[s^v_t; \tilde{V}_t]),
\end{equation}
\begin{equation}
    \small
    p^v_t= Self\_Attn([\hat{s}^v_t; \hat{V}_t]),
\end{equation}
where $s^v_t$ is the new state representation considering visual information (see ``State-V'' in Figure~\ref{fig:model architecture}). $\hat{X^v}, \hat{s}^v_t, \hat{V}_t$ are updated state, text, vision representations after cross modality attention layers in the vision module.
$p^v_t$ is the attention score between state representation and vision representations.
\\
\textbf{Action Selection}
For each navigable viewpoint, we obtain the self-attention scores from 1) orientation state representation to its orientation representation (orientation module), 2) vision state representation to the vision representation (vision module), 3) state representation to the combined orientation and visual representations (history module). We combine these scores as follows:
\begin{equation}
   \small
   p_t = Softmax(W_a[p^h_t; p^o_t; p^v_t])
\end{equation}
where $w_a$ is the trainable parameter, and $p_t$ denote the action probability which weights different module scores. 

\section{Training and Inference}
We train our model with the mixture of Imitation Learning (IL) and Reinforcement Learning (RL) following~\cite{tan2019learning}. It minimizes the cross-entropy loss of the prediction and ground-truth trajectories by following teacher actions for each navigation step. RL samples an action from the action probability $p^a_t$ and learns from the rewards. The loss function is the following:
\begin{equation}
    \small
    l =  -\sum_t a^{s}_t log(p^a_t) - \lambda \sum_t a^{*}_t log(p^a_t) 
\end{equation}
where $a^{*}_t$ is the teacher action, $a^{s}_t$ is the sampled action from RL, and $\lambda$ is the coefficient to balance two training goals.
During inference, we use the greedy search to select the viewpoint with highest probability and finally generate the trajectory.
\section{Pre-training}
\begin{figure}
    \centering
    \includegraphics[width=1.0\linewidth]{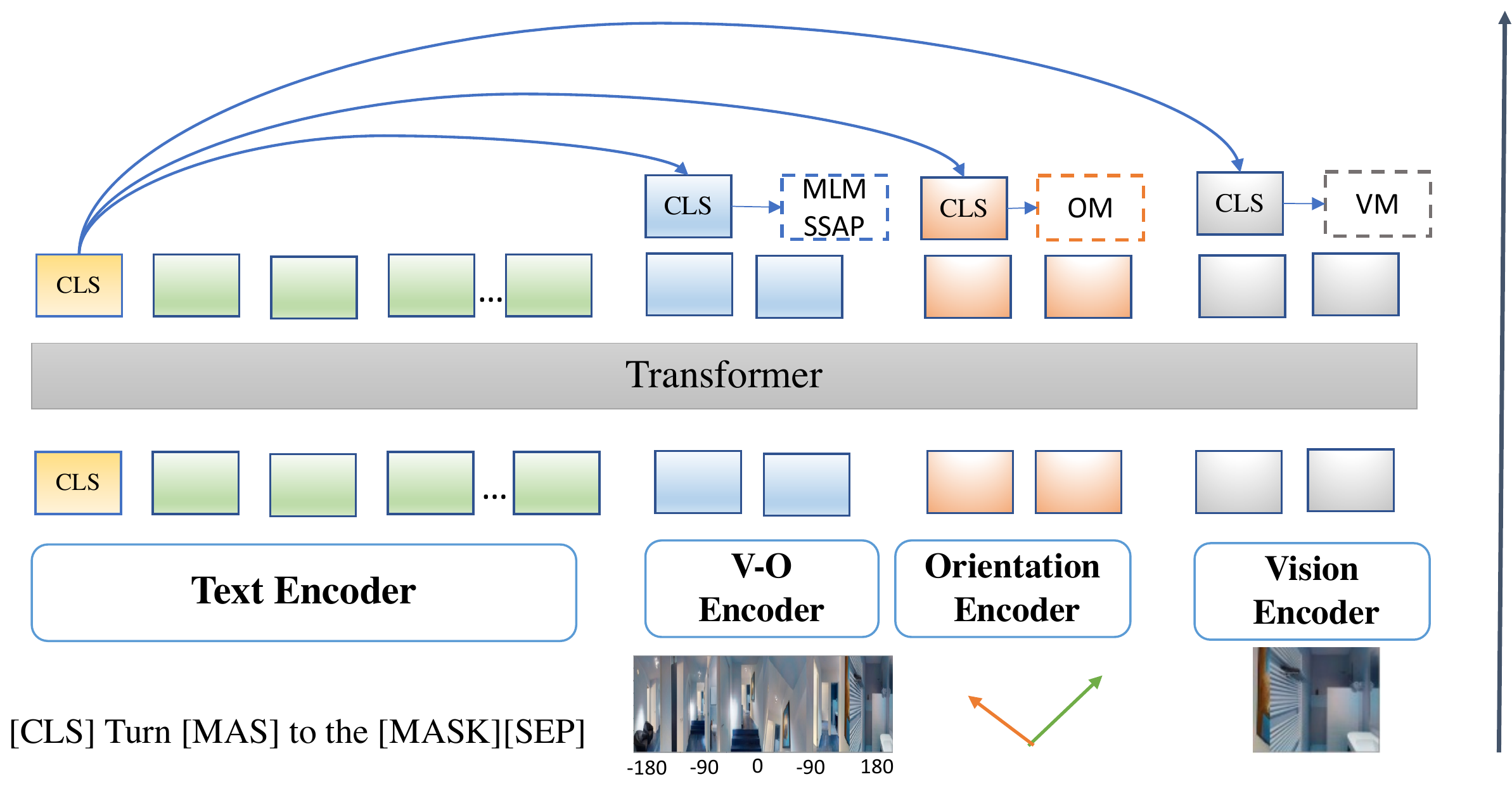}
    \caption{\small Pre-training Model with Specific Pre-training Tasks. V-O is the encoder considering both orientation and vision  representations. MLM: Masked Language Modeling; SSAP: Singe Step Action Prediction; OM: Orientation Matching; VM:Vision Matching.}
    \label{fig:Pretraining Model Architecture}
\end{figure}
We follow the model architecture of PREVALENT~\cite{hao2020towards} to obtain the joint cross representations trained on text-image-action triplets, as shown in Figure~\ref{fig:Pretraining Model Architecture}.
However, the novelty of our pre-training is that we design new tasks named Vision Matching (VM) and Orientation Matching  (OM) to pretrain for the vision module and orientation module designed in our navigation agent, as shown in Figure~\ref{fig:model architecture}. 
Moreover, we improve the existing pre-training tasks of the PREVALENT, Masked Language Modeling (MLM) and Single Step Action Prediction (SSAP), to obtain a more effective initialization of our new architecture. Here, we describe the details of all the pre-training tasks while their effects is described in Section~\ref{Ablation Study of Different Tasks}. 
In the following tasks, we denote each instruction-trajectory pair in training set $D$ as $<w, \tau>$.
\\
\textbf{Masked Language Modeling (MLM)} Different from PREVALENT~\cite{hao2020towards} masking of random tokens, we mask  direction and landmark tokens with 8\% probability and replace them with special token $[MASK]$. The goal is to recover landmark or orientation tokens $w_m$ by reasoning over the surrounding words $w_{\textbackslash m}$, and the orientation and visual observation at the each navigation step. We denote the combination of orientation and vision features of panorama views as $VO_p$. 
Landmark tokens are usually the token related to scene or objects in the visual environment, such as ``
\textit{table}'', ``\textit{sofa}'', and ``\textit{bedroom}''. We extract nouns as landmark tokens based on their pos-tag. The direction tokens usually convey spatial information, such as ``\textit{left}'', ``\textit{right}'', and ``\textit{forward}''.
We obtain direction tokens using a direction dictionary built upon R2R training dataset. The loss of MLM is calculated as follows,
\begin{equation}
    \footnotesize
    \mathcal{L}_{MLM} = -\mathbb{E}_{VO_{p}{\texttildelow} P(\tau), (w, \tau){\texttildelow} D} \log P(w_m|w_{\textbackslash m}, VO_p),
\end{equation}
\\
\textbf{Single Step Action Prediction (SSAP)} PREVALENT~\cite{hao2020towards} selects actions by mapping the $[CLS]$ representations to the 36 classes directly, which may cause the loose connection between cross-modal representations of the viewpoints and the action space. To address this issue, we use the cross attention distribution from the $[CLS]$ representation to the images in the panoramic view to select an action. We use the cross-entropy loss to compute the loss of SSAP, as follows,
\begin{equation}
    \footnotesize
    \mathcal{L}_{SSAP} = -\mathbb{E}_{{OV}_p{\texttildelow} P(\tau), (w, \tau){\texttildelow} D} \log P(a|w_{[CLS]}, VO_p),
\end{equation}
where $a$ is the ground-truth action.
\\
\textbf{Vision Matching (VM)} VM is our novel pre-training specific for initializing our vision module. It predicts whether the current vision information can match with the instruction. 
In this task, to encourage the agent to focus on learning the connection between landmarks in the instruction and the scene objects in the visual environment, we only use the vision representation (i.e. excluding the heading and elevation) of viewpoint as the input, denoted as $v_p$.
We generate the negative samples by replacing the ground-truth images with an images from another environment. We use the output representation of the $[CLS]$ as the joint representation of textual and visual features to feed to a fully connected layer with a sigmoid function. This layer predicts the matching score $s(w, v_p)$. The loss of SSAP is computed as follows,
\begin{equation}
    \small
    \mathcal{L}_{VM} = - \mathbb{E}_{{v}_p{\texttildelow}\tau,(w,\tau)\texttildelow{D}} [y\log P + (1-y) \log P)],
\end{equation}
where $P = s(w, v_p)$, and $y \in \{0,1\}$ indicates whether the sampled viewpoint-instruction pair is matching.
\\
\textbf{Orientation Matching (OM)} Our second novel pre-training task is designed to learn the orientation representations. We propose to predict the current orientation based on the instruction and the initial orientation. 
As described before, the orientation feature $O_p$ is the combination of the heading $\alpha$ and elevation $\beta$. 
We use the output representation of $[CLS]$ as the joint representation of instruction and  orientation. Then we feed this to a fully connected layer to predict 4-bits of orientation features. The loss of OM is computed as follows,
\begin{equation}
    \small
    \mathcal{L}_{OM} = - \mathbb{E}_{{o_p}{\texttildelow}\tau,(w,\tau)\texttildelow{D}}\log p(O'|w_{[CLS]}, O_p),
\end{equation}
where $O'$ is the ground-truth orientation feature.
The full pre-training objective is
\begin{equation}
    \small
     \mathcal{L}_{pre-train} =  \mathcal{L}_{MLM} + \mathcal{L}_{SSAP} +  \mathcal{L}_{VM} +  \mathcal{L}_{OM}.
\end{equation}

\section{Experiments}
\begin{table*}[t]
\small
    \begin{center}
    \begin{tabular}{c c c c c c c c c c c}
    \hline
      &  & \multicolumn{3}{c}{\textbf{Val seen}} & \multicolumn{3}{c}{\textbf{Val Unseen}} & \multicolumn{3}{c}{\textbf{Test(Unseen)}}\\
    \hline
        & Method & \textbf{NE} $\downarrow$ & \textbf{SR} $\uparrow$ & \textbf{SPL}$\uparrow$ & \textbf{NE} $\downarrow$ &  \textbf{SR} $\uparrow$ & \textbf{SPL}$\uparrow$ & \textbf{NE} $\downarrow$&
         \textbf{SR} $\uparrow$& \textbf{SPL} $\uparrow$\\
    \hline
    $1$ & Speaker-Follower \cite{fried2018speaker} & $3.36$ &  $0.66$ &  - & $6.62$ & $0.35$  &  - & $6.62$ & $0.35$ & $0.28$\\
    $2$ & Env-Drop~\cite{tan2019learning}  & $3.99$ & $0.62$ & $0.59$ & $5.22$  & $0.47$ & $0.43$ & $5.23$ & $0.51$ & $0.47$ \\
    $3$ & OAAM~\cite{qi2020object}~ & - & $0.65$ & $0.62$ & - & $0.54$ & $0.50$ & $5.30$ &  $0.53$ & $0.50$ \\
    $4$ & RelGraph~\cite{hong2020language}~& $3.47$ &  $0.67$ & $0.65$ & $4.73$ & $0.57$ & $0.53$ & $4.75$ & $0.55$ & $0.52$ \\
    $5$ & NvEM~\cite{an2021neighbor} & $3.44$ & $0.69$ & $0.65$ & $4.27$ & $0.60$ & $0.55$ & $4.37$ & $0.58$ & $0.54$ \\
    \hline
    $6$ & PRESS~\cite{li2019robust} & $4.39$ & $0.58$ & $0.55$ & $5.28$ & $0.49$ & $0.45$ & $5.49$ & $0.49$ & $0.45$ \\
    $7$ & PREVALENT~\cite{hao2020towards} & $3.67$ & $0.69$ & $0.65$ &  $4.71$ &  $0.58$ & $0.53$ & $5.30$ & $0.54$ & $0.51$\\
    $8$ & AirBERT~\cite{guhur2021airbert} & $2.68$ & $0.75$ & $0.70$ & $4.01$ & $0.62$ & $0.56$ & $4.13$ & $0.62$ & $0.57$\\
    $9$ & RecBERT~\cite{hong2021vln} & $2.90$ & $0.72$ & $0.68$ & $3.93$ & $0.63$ & $0.57$ & $4.09$ & $\textbf{0.63}$ & $0.57$\\
    $10$ & HAMT~\cite{chen2021history} & - & $0.69$ & $0.65$ & - & $0.64$ & $0.58$ & - & - & - \\
    \hline
    11 & RecBERT* & 2.99 & 0.71 & 0.66 & 4.03 & 0.61 & 0.56 & 4.35 & 0.61 & 0.57 \\
    12 & Our pretrain + RecBERT &  2.90 & 0.74 & 0.69 & 3.75 & 0.63 & 0.58 & 4.20 & \textbf{0.63} & 0.57\\
    13 & Our pretrain + LOViS~(our model) & \textbf{2.40} &\textbf{0.77} & \textbf{0.72} & \textbf{3.71} & \textbf{0.65} & \textbf{0.59} & \textbf{4.07} & \textbf{0.63} & \textbf{0.58}\\
    \hline
    \end{tabular}
    \end{center}
    \vspace{-4mm}
    \caption{\small Experimental Results Comparing with Baseline Models on R2R Benchmarks in a single-run setting. The best results are in bold font. * denotes our reproduced R2R results.}
    \label{Experimental Result comparing with other models.}
       \vspace{-2mm}
\end{table*}

\begin{table*}[!t]
\small
\renewcommand{\tabcolsep}{0.35em}
    \begin{center}
    \begin{tabular}{ c c c c c c c c c c c c c}
    \hline
    & \multicolumn{6}{c}{\textbf{Val Seen}} & \multicolumn{6}{c}{\textbf{Val Unseen}} \\
    \hline
    Method & \textbf{NE}$\uparrow$ & \textbf{SR}$\uparrow$ &  \textbf{SPL}$\uparrow$ & \textbf{CLS}$\uparrow$ &\textbf{nDTW}$\uparrow$ & \textbf{sDTW}$\uparrow$ & \textbf{NE}$\downarrow$ & \textbf{SR}$\uparrow$ & \textbf{SPL}$\uparrow$ & \textbf{CLS}$\uparrow$ & \textbf{nDTW}$\uparrow$ & \textbf{sDTW}$\uparrow$  \\
    \hline
    EnvDrop*~\cite{tan2019learning}  & - & $0.52$ & $0.41$ & $0.53$ & - & $0.27$ & - & $0.29$ & $0.18$ & $0.34$ & - & $0.09$  \\
    OAAM~\cite{qi2020object} & - & $0.56$ & $0.49$ & $0.54$ & - & $0.32$ & - & $0.29$ & $0.18$ & $0.34$ & - & $0.11$\\
    NvEM~\cite{an2021neighbor}  & $5.38$ & $0.54$ & $0.47$ & $0.51$ & $0.48$ & $0.35$ & $6.80$ & $0.38$ & $0.28$ & $0.41$ & $0.36$ & $0.20$\\
    \hline
    RecBERT*~\cite{hong2021vln}  & $4.82$ & $0.56$ & $0.46$ & $0.50$ & $0.56$ & $0.38$ & $6.48$ & $0.43$ & $0.32$ & $0.41$ & $0.42$ & $0.21$\\
    LOViS~(our model) & $\textbf{4.16}$ & $\textbf{0.67}$ & $\textbf{0.58}$ & $\textbf{0.56}$ & $\textbf{0.58}$ & $\textbf{0.43}$ & $\textbf{6.07}$ & $\textbf{0.45}$ & $\textbf{0.35}$ & $\textbf{0.45}$ & $\textbf{0.43}$ & $\textbf{0.23}$\\
    \hline
    \end{tabular}
    \end{center}
    \vspace{-3mm}
    \caption{\small Experimental Results for comparing LOViS with the Baseline Models on R4R dataset in a single-run setting. The best results are in bold font. * denotes our reproduced R4R results.\vspace{-3mm}}
    \label{Experiment Results for R4R}
\end{table*}

\subsection{Dataset} 
Two VLN datasets are used in evaluation: R2R~\cite{anderson2018vision} and R4R~\cite{jain2019stay}.\\
\textbf{R2R} builds upon the Matterport3D dataset. 
This dataset has 7198 paths and 21567 instructions with an average length of 29 words. The whole dataset is partitioned into training, seen validation, unseen validation, and unseen test set. The seen set shares the same visual environments with the training set, while unseen sets contain different environments. \\
\textbf{R4R} extends the R2R dataset with longer instructions and trajectories by concatenating two adjacent tail-to-head trajectories in R2R. Different from R2R, the trajectories in R4R are less biased as they are not necessarily the shortest path from the start viewpoint to the destination.

\subsection{Evaluation Metrics} We mainly report three evaluation metrics for R2R. (1) Navigation Error (NE): the mean of the
shortest path distance between the agent’s final position and the goal location.  (2) Success Rate (SR): the percentage of the cases where the predicted final position is close within 3 meters from the goal location. (3) Success rate weighted by normalized inverse Path Length (SPL)~\cite{anderson2018vision}: normalizes Success Rate by trajectory length. It considers both the effectiveness and efficiency of navigation performance.

In terms of the metrics for the R4R benchmark, besides the basic metrics same as R2R, NE, SR, and SPL, R4R includes additional metrics: the Coverage Weighted by Length Score (CLS), the Normalized Dynamic Time Warping and the nDTW weighted by Success Rate (sDTW). In R4R, SR and SPL measure the performance of the navigation, while CLS, nDTW and sDTW measure the fidelity of the predicted paths.

\subsection{Implementation Details}
Please check our code~\footnote{\url{https://github.com/HLR/LOViS}} for the implementation.\\
\textbf{Pre-training} We use 4 GeForce RTX 2080 GPUs for pre-training. The batch size for each GPU is 28, and the training time is around 22 hours. The learning rate is $5e-5$, and the AdamW optimizer is adopted. The language Transformer has nine layers, and the cross-modality Transformer has four layers. The models' parameters are initialized with the weights of PREVALENT~\cite{hao2020towards}.\\
\textbf{Fine-tuning}
We directly adapt different encoders and Transformer layers from our pre-training model to our navigation model.
The navigation model is further trained in 30k iterations with learning rate $1e-5$. The batch size is $28$. The best performance is selected according to the best SPL of the validation unseen set. For R2R, we use the same augmented data as in~\cite{hong2021vln} for a fair comparison. 

\begin{table*}[!t]
\small
\renewcommand{\tabcolsep}{0.35em}
    \begin{center}
    \begin{tabular}{ c l c c c c | c c c c}
    \hline
      & & \multicolumn{4}{c}{Baseline Model} & \multicolumn{4}{c}{LOViS (Our Model)}  \\
    \hline
      & & \multicolumn{2}{c}{Val Seen} & \multicolumn{2}{c}{Val Unseen} & \multicolumn{2}{c}{Val Seen} & \multicolumn{2}{c}{Val Unseen} \\
    \hline
        & Tasks & \textbf{SR}$\uparrow$ & \textbf{SPL}$\uparrow$  & \textbf{SR}$\uparrow$ & \textbf{SPL}$\uparrow$ & \textbf{SR}$\uparrow$ & \textbf{SPL}$\uparrow$  & \textbf{SR}$\uparrow$ & \textbf{SPL}$\uparrow$ \\
    \hline
    $1$ & MLM & $0.712$ & $0.662$ & $0.613$ & $0.562$ & $0.724$ & $0.673$ & $0.621$ & $0.564$\\
    \hline
    $2$ & MLM+SSAP & $0.731$ & $0.675$  & $0.619$ & $0.575$ & $0.747$ & $0.695$ & $0.649$ & $0.585$\\
    \hline
    $3$ & MLM+SSAP+VM & $0.737$ & $0.683$ & $0.622$ & $0.577$& $0.755$ & $0.711$ & $0.637$ & $0.581$ \\
    \hline
    $4$ & MLM+SSAP+OM & $0.730$ & $0.672$ & $0.617$ & $0.574$ & $0.766$ & $0.724$ & $0.629$ & $0.579$\\
    \hline
    $5$ & MLM+SSAP+VM+OM & $\textbf{0.743}$ & $\textbf{0.691}$ & $\textbf{0.632}$ & $\textbf{0.583}$ & $\textbf{0.774}$ & $\textbf{0.722}$ & $\textbf{0.653}$ & $\textbf{0.592}$\\
    \hline
    \end{tabular}
    \end{center}
    \vspace{-4mm}
    \caption{\small Ablation Study for Different Tasks of Pre-training on the Baseline and LOViS.}
    \label{Ablation Study for Different Tasks in Pre-training.}
\end{table*}

\subsection{Comparisons with SoTA}
Table~\ref{Experimental Result comparing with other models.} shows the experimental results of different VLN methods on R2R benchmarks in a single-run setting. In this table, row\#1 to row\#5 show the results of the LSTM-based navigation agents. From row\#6 to row\#10 are Transformer-based navigation agents that largely have improved the performance of the LSTM-based agents.
PREVALENT~\cite{hao2020towards} pre-trains the cross-modal representations with text-image-action triplets and replaces the encoder of Env-Drop~\cite{tan2019learning} to improve its performance. AirBERT~\cite{guhur2021airbert} is one of the SOTA methods that train a model on a large scale and diverse in-domain detests.
RecBERT\cite{hong2021vln}~, our baseline, is also a SOTA method that uses the attention distribution of the history information on navigation candidates to determine the next action.
Row\#10 is their own reported results in their paper, and row\#11 shows our best reproduced results which is consistent with the reported results in~\cite{liu2021vision}.
Row\#12 and row\#13 are the performance of our LOViS model. We first show the effectiveness of our pre-training on the baseline model. Our pre-training setting can improve the SR and SPL of baseline by about 2\% in the unseen validation environment. Moreover, we further improve the performance of the baseline with our designed navigation model and the pre-training setting. The improvement is about  3\% of SR and SPL in the seen environment and 2\% of SR in the unseen validation and test environment. This result indicates our pre-training tasks are more suitable for our designed navigation model. We also obtain a lower NE showing that our agent navigates closer to the destination.
For HAMT~\cite{chen2021history}, we report their results with ResNet-152 as the vision encoder for a fair comparison.
Among those methods, only OAAM~\cite{qi2020object} and NvEM~\cite{an2021neighbor} consider the semantics of spatial information and visual perception, but their results have a large performance gap compared to our Transformer-based navigation model. 

Table~\ref{Experiment Results for R4R} shows the performance of various models on R4R benchmark in a single-run setting. Same as R2R, we can better perform in all evaluation metrics. Compared to the our reproduced results of the RecBERT~\cite{hong2020language}, we can improve 4\% of CLS, 1\% of nDTW, and 2\% of sDTW in the unseen validation environment, which indicates the significantly better fidelity of our model.

\subsection{Ablation Study}
\begin{table}[!t]
\small
\renewcommand{\tabcolsep}{0.35em}
    \begin{center}
    \begin{tabular}{ c l c c c c}
    \hline
      & & \multicolumn{2}{c}{Val Seen} & \multicolumn{2}{c}{Val Unseen} \\
    \hline
        & Modules & \textbf{SR}$\uparrow$ & \textbf{SPL}$\uparrow$ & \textbf{SR}$\uparrow$ & \textbf{SPL}$\uparrow$  \\
    \hline
    $1$ & H & $0.743$ & $0.691$ & $0.632$ & $0.583$ \\
    \hline
    $2$ & H+O &  $0.756$ & $0.712$ & $0.629$ & $0.576$\\
    \hline
    $3$ & H+V  & $0.762$ & $0.718$ & $0.642$ & $0.588$\\
    \hline
    $4$ & H+O+V & $\textbf{0.774}$ & $\textbf{0.722}$ & 
    $\textbf{0.653}$ & $\textbf{0.592}$ \\
    \hline
    \end{tabular}
    \end{center}
    \vspace{-3mm}
    \caption{\small Ablation Study for Different Modules in 
    Model. H: History Module; O: Orientation Module; V: Vision Module.}
    \label{Module Ablation Study}
     \vspace{-5mm}
\end{table}
\subsubsection{Ablation Study of Different Tasks}
\label{Ablation Study of Different Tasks}
In Table~\ref{Experimental Result comparing with other models.}, we already observe that our pre-training strategy improved the RecBERT baseline. In Table~\ref{Ablation Study for Different Tasks in Pre-training.}, we show the influence of each pre-training task on both RecBERT and LOViS. For RecBERT baseline model, SSAP shows about 2\% of improvement on both seen and unseen environments. Although the tasks of VM and OM independently do not change the performance of MLM+SSAP, the combination of two tasks improves the performance by about 1\%. The same phenomenon happens in LOViS. SSAP improves the performance by a large margin. Although VM and OM do not show significant improvement when used separately in the unseen environment, they improve both SR and SPL in the seen environment. The combination of VM and OM improves the performance significantly, especially in the seen environment.

\subsubsection{Ablation Study of Different Modules}
Table~\ref{Module Ablation Study} shows the ablation of different modules. 
Based on row\#1, the history module with our pre-training strategy has already improved its performance. Comparing row\#2 and row\#3, we can see that vision module affects the results more than the Orientation module. The combination of two modules with our pre-training strategy achieves the best performance (row\#4). This indicates that our designed explicit modules can assist the agent  in choosing the correct action based on different information.

\subsection{Qualitative Example}
\begin{figure}
    \centering
    \includegraphics[width=1.0\linewidth]{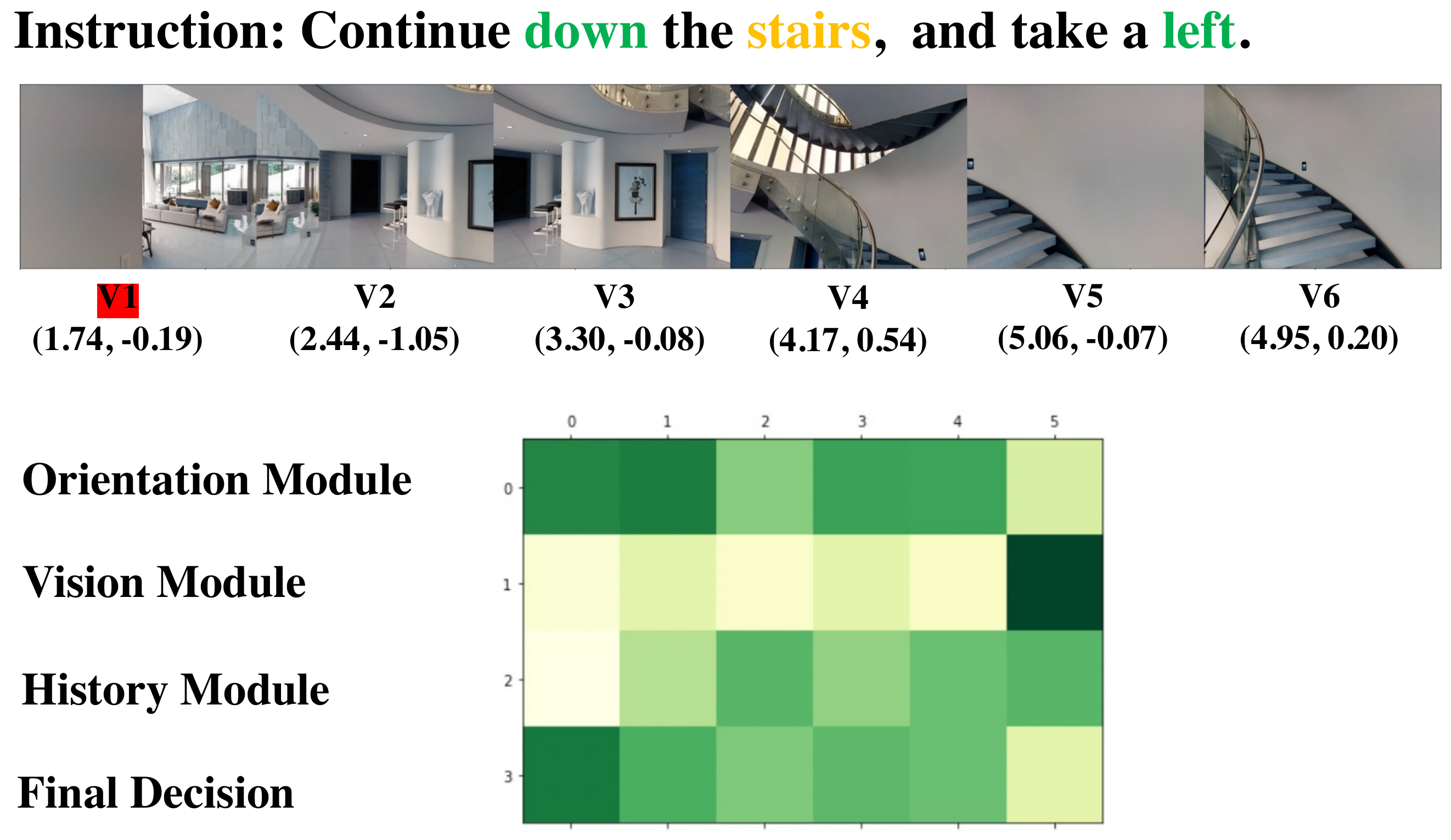}
    \caption{\small \textbf{Qualitative Example.} The ground-truth viewpoint is v1. The word ``\textit{down}'' and ``\textit{left}'' are the orientation signals. The word ``\textit{stairs}'' is the vision signal. The attention map shows the score of different candidate viewpoints in each module. The darker color means the higher score. The numbers below each viewpoint show the orientation information with the format of <relative heading, relative elevation>. The lower value of each number means the orientation is more towards left and down respectively.}
    \vspace{-5mm}
    \label{fig:qualitative example}
\end{figure}

Figure~\ref{fig:qualitative example} shows a qualitative example that demonstrates the performance of each module of LOViS navigation agent. It is evident that the \textit{orientation module} gives a higher score to the viewpoints that are left, and their elevation is down. The \textit{vision module} gives a higher score to the viewpoints that ``stairs'' can be seen. The history module also gives a relatively higher score to the viewpoints on the right side. The final decision is $v1$ with its weights of $[0.02, -0.03, -0.04]$ to the three modules. 
The example shows that our designed orientation and vision modules can attend to the viewpoint with the corresponding information.

\section{Conclusion}

The main idea of this paper is to design explicit vision and orientation modules in the neural architecture of a navigating agent. These modules can effectively learn to ground the landmark mentions and spatial information related to the orientation of the agent expressed in the natural language instruction into the visual environment. To make the designed modules more effective, we design new pre-training tasks accordingly to equip the agent with spatial reasoning and visual perception abilities before navigation. We evaluate our model on R2R and R4R datasets and achieve state-of-the-art results. Our ablation study shows the effectiveness of our designed modules and pre-training tasks.
\section{Acknowledgement}
This project is supported by National Science Foundation (NSF) CAREER award 2028626 and partially supported by the Office of Naval Research
(ONR) grant N00014-20-1-2005. Any opinions,
findings, and conclusions or recommendations expressed in this material are those of the authors and
do not necessarily reflect the views of the National
Science Foundation nor the Office of Naval Research. We thank all reviewers for their thoughtful
comments and suggestions.

\bibliography{anthology,custom}
\bibliographystyle{acl_natbib}

\appendix



\end{document}